\documentclass[10pt,twoside]{article}
\usepackage{float}
\usepackage{times}
\usepackage[utf8]{inputenc}
\usepackage[T1]{fontenc}
\usepackage{graphicx}
\usepackage[table,xcdraw]{xcolor}
\usepackage[normalem]{ulem}
\useunder{\uline}{\ul}{}
\usepackage{multirow}
\usepackage[most]{tcolorbox}
\usepackage{xcolor}
\usepackage{lmodern}
\usepackage{enumitem} 
\usepackage{amssymb}
\usepackage{pifont}
\usepackage{titlesec}

\titlespacing{\section}{0pt}{0.8\baselineskip}{0.5\baselineskip}
\titlespacing{\subsection}{0pt}{0.8\baselineskip}{0.5\baselineskip}
\titlespacing{\subsubsection}{0pt}{0.8\baselineskip}{0.5\baselineskip}
\titlespacing{\paragraph}{0pt}{0.8\baselineskip}{0.5\baselineskip}

\usepackage{caption}
\usepackage{siunitx}
\usepackage{hyperref}
\usepackage{cleveref}
\usepackage{coria-taln2025}
\usepackage[french]{babel} 

\title{LLM, Au Rapport !\\ Extraction d'Informations Médicales entre Prompting, Fine-tuning et Post-correction}

\author{Ikram Belmadani\up{2}\quad Parisa Nazari Hashemi\up{1, 4}\thanks{Contribution équivalente.}\quad Thomas Sebbag\up{1, 3}\footnotemark[1]\\Benoit Favre\up{2}\quad Guillaume Fortier\up{4}\quad Solen Quiniou\up{1}\quad Emmanuel Morin\up{1}\quad Richard Dufour\up{1}\\
  {\small
    (1) Nantes Université, École Centrale Nantes, CNRS, LS2N, UMR 6004, F-44000 Nantes, France \\ 
    (2) Aix-Marseille Université, CNRS, LIS UMR 7020, 13000, Marseille, France \\ 
    (3) Explore, Carquefou, France, (4) Inetum, 93400 Saint-Ouen-sur-Seine, France \\ 
    \texttt{
      \up{1}prenom.nom@univ-nantes.fr, \up{2}prenom.nom@univ-amu.fr, \up{4}prenom.nom@inetum.com \\ 
}}}

\begin{document}
\maketitle

\resume{
Ce travail présente notre participation au défi EvalLLM 2025 portant sur la reconnaissance d'entités nommées (REN) biomédicales et l'extraction d'événements sanitaires en français en contexte {\it few-shot}. Pour la REN, nous proposons trois approches intégrant des très grands modèles de langue (LLM), guide d'annotation, données synthétiques, et post-traitement : (1) apprentissage par contexte (ICL) avec GPT-4.1,  intégrant une sélection automatique de 10 exemples et le résumé du guide d'annotation dans le prompt, (2) système universel de REN (GLiNER) affiné sur un corpus synthétique, puis vérifié par LLM en post-traitement, et (3) LLM ouvert (LLaMA-3.1-8B-Instruct) affiné sur le même corpus synthétique. L'extraction d'événements exploite la même stratégie ICL avec GPT-4.1, en réutilisant le guide résumé dans le prompt. Les résultats montrent que GPT-4.1 domine, avec une macro-F1 de 61,53~\% en REN et 15,02~\% en extraction d'événements, soulignant l'importance d'un prompting bien formé pour maximiser les performances en très faibles ressources.
}

\abstract{LLM, Reporting In! Medical Information Extraction Across Prompting, Fine-tuning and Post-correction}{
This work presents our participation in the EvalLLM 2025 challenge on biomedical Named Entity Recognition (NER) and health event extraction in French (few-shot setting). For NER, we propose three approaches combining large language models (LLMs), annotation guidelines, synthetic data, and post-processing: (1) in-context learning (ICL) with GPT-4.1, incorporating automatic selection of 10 examples and a summary of the annotation guidelines into the prompt, (2) the universal NER system GLiNER, fine-tuned on a synthetic corpus and then verified by an LLM in post-processing, and (3) the open LLM LLaMA-3.1-8B-Instruct, fine-tuned on the same synthetic corpus. Event extraction uses the same ICL strategy with GPT-4.1, reusing the guideline summary in the prompt. Results show GPT-4.1 leads with a macro-F1 of 61.53\% for NER and 15.02\% for event extraction, highlighting the importance of well-crafted prompting to maximize performance in very low-resource scenarios.
}

\motsClefs
  {LLM, Reconnaissance entités nommées, données synthétiques, correction}
  {LLM, Named Entity Recognition, Synthetic Data, Correction}

\section{Introduction}
L'extraction d'informations, reposant sur différentes techniques issues du Traitement Automatique des Langues (TAL), englobe une diversité de tâches présentant des verrous importants~\citep{niklaus-etal-2018-survey}. Parmi ces tâches, l'extraction d'événements s'avère particulièrement complexe, en raison de la nature souvent implicite qui caractérise l'expression d'un événement dans un texte. Cette difficulté est renforcée par l'inter-dépendance avec une tâche préliminaire essentielle : la reconnaissance d'entités nommées (REN), constituant l'événement en tant que tel, ou faisant partie du contexte qui permettra de le définir. 

L'atelier EvalLLM2025, proposé dans le cadre de la conférence CORIA-TALN 2025, organise un challenge d'évaluation par la tâche dans le domaine de la santé en français. Les données sont issues de documents journalistiques utilisés pour la veille sanitaire, dans un contexte {\it few-shot} où peu de données annotées sont disponibles pour caractériser chaque type d'entités ou d'événements.
Dans ce contexte, avec l'émergence des très grands modèles de langue (LLM) et d'approches par \textit{few-shot learning} (FSL), il est devenu possible, sur la base de quelques exemples, d'apprendre une nouvelle tâche, ce qui est particulièrement intéressant dans un cadre où les données d'entraînement disponibles sont très limitées. L'application de cette méthode à la REN~\citep{ma-etal-2022-template} ou à l'extraction d'événements~\citep{ma-etal-2023-shot, yue-etal-2023-zero} a été la base de nombreux travaux. Cependant, bien que le FSL permette d'obtenir des résultats rapidement, l'approche n'est souvent pas la plus optimale, les performances atteignant un plafond de verre selon~\citet{zhang-etal-2025-survey}.


Dans cet article, nous décrivons les méthodes proposées pour la campagne EvalLLM pour la REN et l'extraction d'événements, intégrant plusieurs approches par LLM avec fine-tuning au travers de données synthétiques, un prompting via un résumé du guide d'annotation et une sélection d'exemples ({\it few-shot}), ou encore une vérification en post-traitement au moyen d'un LLM. Le code source des différentes approches ainsi que les ressources générées sont disponibles en ligne\footnote{\url{https://github.com/ikram28/EvalLLM2025.git}} afin de permettre la réplicabilité des résultats.





\section{Méthodologie}

Cette section présente les modèles initiaux sur lesquels nous nous sommes appuyés (Sous-section~\ref{sec:mod_base}) puis décrit les différents modules et techniques que nous avons explorés (Sous-section~\ref{sec:tech}). Ces modules ont ensuite été combinés sous la forme de pipelines dans différentes configurations expérimentales qui seront présentées dans la Section~\ref{sec3}. 

\subsection{Modèles de base}
\label{sec:mod_base}

Pour nos approches, nous avons sélectionné deux LLM génératifs — le modèle propriétaire GPT-4.1 et le modèle open source LLaMA-3.1 — ainsi que GLiNER, un outil spécialisé pour la tâche de NER fondé sur l'architecture BERT~\citep{devlin2019bert}.

\paragraph{GPT-4.1} ~est un LLM multimodal capable de traiter du texte et des images en entrée pour générer du texte en sortie~\citep{openai_gpt-4_2024}. Il a démontré des capacités avancées dans diverses tâches de génération, notamment la rédaction de descriptions, récits, poèmes, publicités ou code~\citep{10.5555/3540261.3541003, 10.5555/3600270.3602654, zeng2022x}. Ces performances reposent notamment sur l'architecture Transformer~\citep{vaswani2017attention}, des ressources d'entraînement massives, et une forte implication humaine dans la formulation et la diversification des prompts~\citep{zeng_what_2024}.
\paragraph{LLaMA-3.1-8B-Instruct} ~est un LLM généraliste\footnote{\url{https://huggingface.co/meta-llama/Llama-3.1-8B-Instruct}} conçu pour traiter un large éventail de tâches en TAL. La variante 8B-Instruct correspond à la version la plus compacte de la famille LLaMA 3~\citep{grattafiori_llama_2024}, affinée par instruction afin d'améliorer sa capacité à suivre des consignes formulées en langage naturel.
\paragraph{GLiNER-biomed}~\citep{yazdani_gliner-biomed_2025} est un modèle de REN s'appuyant sur l'architecture BERT~\citep{devlin2019bert}. Ce modèle particulier a été un dérivé du modèle généraliste GLiNER~\citep{zaratiana-etal-2024-gliner} puis spécialisé dans le domaine médical. La principale innovation de GLiNER consiste à formuler la REN comme un problème d'appariement dans un encodeur unique qui représente conjointement le texte et les étiquettes dans un contexte {\it zero-shot}. GLiNER-biomed a été entraîné avec deux jeux de données : un jeu de pré-entraînement synthétique dédié à la REN dans des textes biomédicaux, puis un jeu de post-entraînement sur des données généralistes afin de conserver la capacité du modèle à faire de l'extraction en {\it zero-shot}.

\subsection{Techniques employées}
\label{sec:tech}

\subsubsection{Augmentation de données}
\label{sec2.2.1}

\begin{figure}[htbp] 
\begin{center} 
\includegraphics[width=0.9\linewidth]{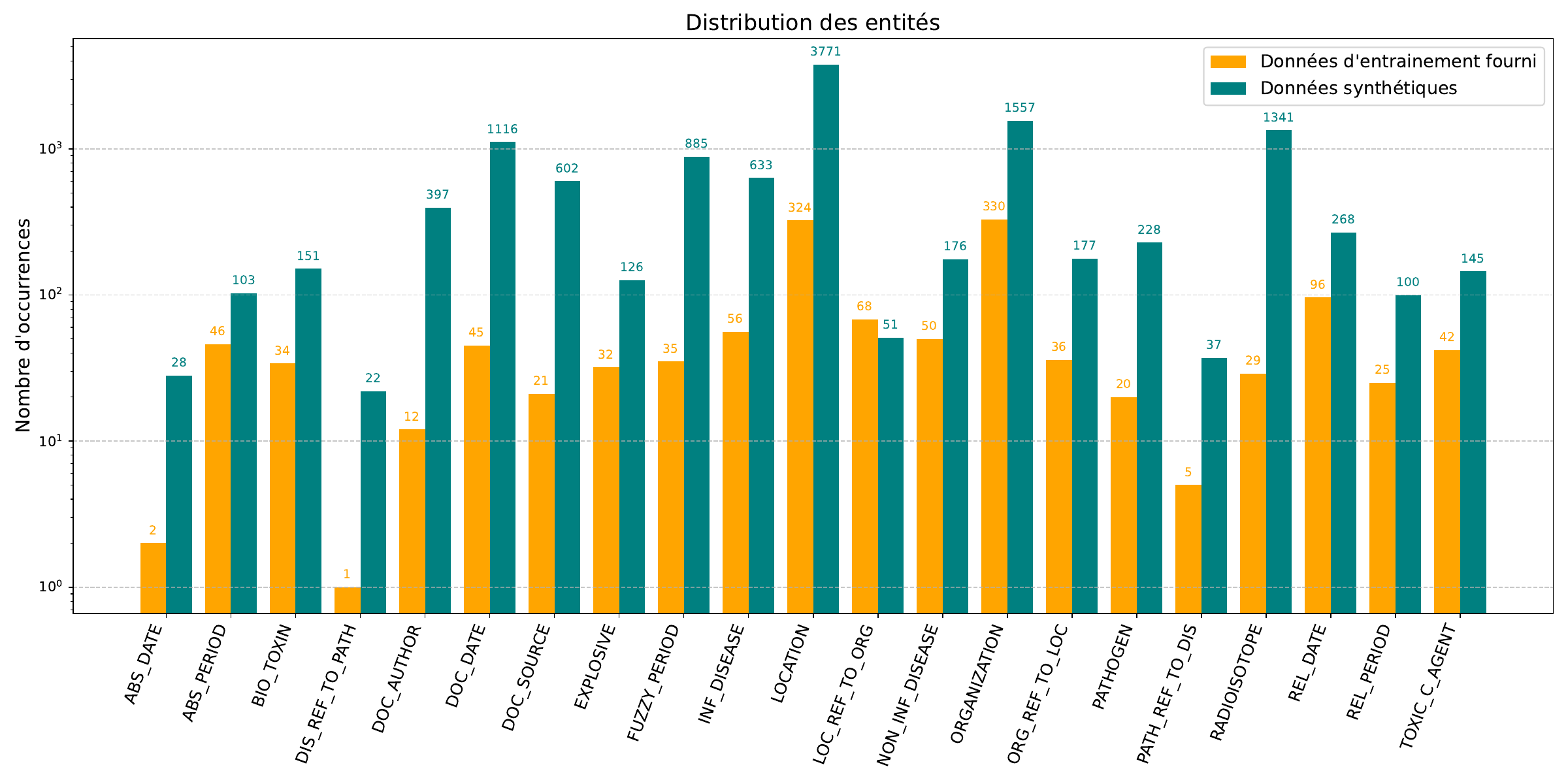}
\end{center} 
\caption{Répartition des entités dans les jeux de données (entraînement et synthétique).}
\label{fig:labels_dist} \
\end{figure}

L'ensemble d'entraînement comprenant seulement 40 documents annotés, nous avons considéré que ce volume n'était pas idéal pour un affinage exploitant les pleines capacités des modèles. Nous avons donc utilisé une stratégie d'augmentation des données avec GPT-4.1. À partir de chaque exemple du jeu d'entraînement, nous avons généré 40 variantes annotées. Pour accroître la diversité des exemples synthétiques, nous avons modulé le paramètre de température du modèle lors de la génération. Un post-traitement automatique a été appliqué pour corriger les décalages dans les positions des spans des entités et éliminer les exemples mal formatés. Au total, cette approche a permis de produire 1~748 exemples synthétiques annotés. La distribution détaillée des entités par type est illustrée dans la Figure~\ref{fig:labels_dist}. Les données synthétiques sont disponibles en ligne\footnote{\url{https://huggingface.co/datasets/ik-ram28/synthetic-NER-dataset}}.

\subsubsection{Fine-tuning supervisé}
Deux procédures de fine-tuning ont été appliquées selon l'architecture. Pour \textbf{GLiNER}, les étiquettes ont été converties en descriptions textuelles (ex. : "Maladie infectieuse" plutôt que "INF\_DISEASE") afin d'exploiter les embeddings de labels. Pour \textbf{LLaMA}, un affinage via LoRA~\citep{hu2021lora} a été réalisé, permettant d'adapter le modèle au domaine biomédical tout en préservant ses capacités générales.

\subsubsection{Prompting et ICL}
\label{sec2.2.3}
Nous avons utilisé une approche d'apprentissage par contexte (ICL) qui exploite de manière optimale les ressources fournies par le défi. Les exemples {\it few-shot} sont sélectionnés automatiquement par similarité cosinus avec le texte à traiter, cherchant ainsi une pertinence contextuelle. Le guide d'annotation fourni par les organisateurs est intégré sous forme de résumé structuré dans le prompt système, permettant ainsi au modèle de bénéficier des définitions précises des entités et des contraintes spécifiques du domaine biomédical. Diverses formulations du prompt ont été évaluées, incluant la version brute du guide, des reformulations manuelles, ainsi que des versions optimisées automatiquement à l'aide de l'outil ChatGPT Prompt Engineer\footnote{\url{https://chatgpt.com/g/g-5XtVuRE8Y-prompt-engineer}}. Ce processus d'optimisation vise à maximiser la clarté des instructions tout en préservant l'information technique essentielle.

\subsubsection{Post-vérification par LLM}
\label{sec2.2.4}
Nous avons développé un module de post-vérification par LLM afin d'améliorer la couverture des entités extraites par le modèle spécialisé. Nos observations préliminaires ont montré qu'après fine-tuning, GLiNER obtenait une précision élevée mais souffrait d'un rappel insuffisant, ce qui limitait la performance globale. Pour remédier à ce déséquilibre, une étape complémentaire de vérification a été introduite : le LLM reçoit les entités extraites par un modèle ainsi que le texte source, et est chargé de valider les prédictions initiales tout en identifiant les entités potentiellement manquantes. Cette combinaison vise à tirer parti de la précision d'un premier modèle tout en exploitant les capacités de généralisation d'un LLM.

\subsubsection{Gestion des formats et nettoyage}
\label{sec2.2.5}
Afin de garantir la cohérence entre les sorties des modèles et le format attendu par les outils d'évaluation, plusieurs modules de nettoyage ont été mis en place.
\paragraph{Conversion au format XML} ~Pour les approches basées sur des LLM, les entités sont générées sous forme de balises XML directement insérées dans le texte, par exemple : \textit{<RADIOISOTOPE>uranium 238</RADIOISOTOPE>}. Ce format structuré a été adopté à la suite de nos premières expérimentations, qui ont montré que la génération directe des positions de début et de fin des entités (spans) induisait fréquemment des erreurs d'alignement textuel. L'utilisation de balises XML permet ainsi une extraction automatique plus fiable et robuste des entités à partir du texte généré.
\paragraph{Alignement automatique} ~Étant donné que les LLM peuvent introduire de légères modifications dans la structure du texte (espaces superflus, ponctuation différente, etc.), un alignement entre le texte source et la sortie du modèle est effectué afin de retrouver les positions exactes de chaque entité dans le texte original. Une vérification finale est ensuite appliquée pour s'assurer que les indices de début et de fin des entités extraites correspondent bien aux occurrences réelles dans le texte. En cas d'incohérence, les spans sont ajustés automatiquement ou ignorés si la correction n'est pas jugée fiable. 

\section{Expérimentations}
\label{sec3}

\begin{figure}[ht] 
\begin{center} 
\includegraphics[width=0.9\linewidth]{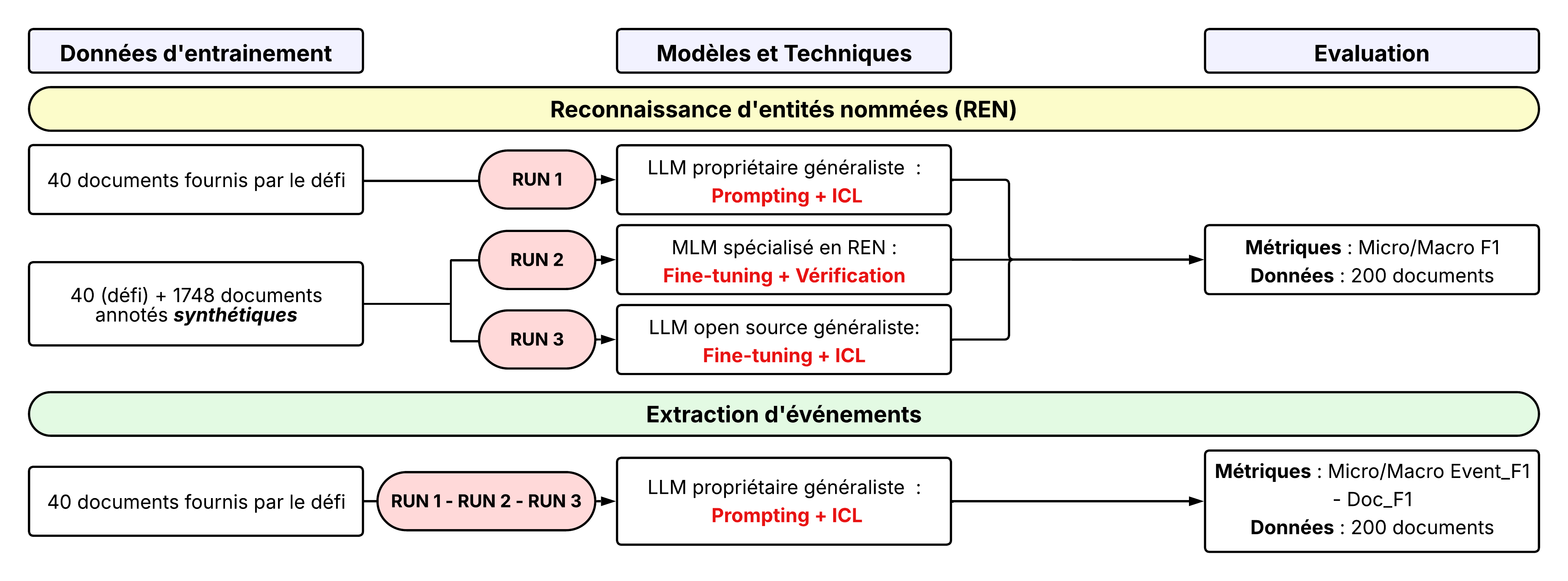}
\end{center} 
\caption{Pipeline pour chaque run soumis à la campagne EvalLLM.}
\label{fig:pipeline} \
\end{figure}

Afin de mener à bien nos expérimentations, nous proposons trois configurations expérimentales (runs) distinctes, chacune évaluée selon le même protocole sur les données fournies par le challenge. Ce dernier comporte deux sous-tâches : la REN couvrant 21 types d'entités, et l'extraction d'événements sanitaires. Pour la REN, nous avons testé trois approches différentes (une par run), tandis que l'extraction d'événements repose sur la même approche dans les trois runs. Le développement et l'optimisation des hyper-paramètres ont été réalisés sur l'ensemble d'entraînement fourni par le challenge, utilisé ici comme corpus de développement, afin de déterminer les meilleures configurations avant la phase de test. La Figure~\ref{fig:pipeline} présente les différents pipelines utilisés pour chaque run.

\label{sec2}

\subsection{Tâche de reconnaissance d'entités nommées (REN)}
\subsubsection{Run 1 : Approche GPT-4.1 en apprentissage par contexte}
Ce run repose sur l'utilisation directe de GPT-4.1 sans phase de fine-tuning, avec pour objectif de tirer pleinement parti du guide d'annotation et des exemples fournis. Un prompt optimisé est conçu (Annexe \ref{ann:A}), intégrant un résumé structuré du guide ainsi qu'un contexte {\it few-shot} (10 exemples), conformément à la stratégie décrite en Section~\ref{sec2.2.3}. Les entités sont générées au format XML, puis un alignement automatique est appliqué pour récupérer les spans corrects, selon la méthode présentée en Section~\ref{sec2.2.5}. Nous avons choisi cette version de GPT-4 sur la base de résultats préliminaires favorables en termes de performance et de coût.

\subsubsection{Run 2 : Approche GLiNER avec post-vérification par LLM}
Dans cette configuration, nous utilisons un modèle mixte que nous appelons EvalLLM-GLiNER, fondé sur gliner-biomed-large-v1.0 et affiné sur notre corpus synthétique (Section~\ref{sec2.2.1}). L'entraînement a été réalisé sur trois époques, en conservant le meilleur {\it checkpoint} ({\it validation loss} minimale à 2,85 époques). Les hyper-paramètres utilisés sont : taux d'apprentissage de 1e-5, {\it weight decay} de 0.01, {\it scheduler} de type cosinus avec 10~\% de {\it warm-up}, et une taille de {\it batch} de 8.
Une fois les entités extraites, GPT-4.1 est utilisé pour la vérification, conformément à la méthode décrite en Section~\ref{sec2.2.4}. Le prompt utilisé est détaillé en Annexe~\ref{annC}.
\subsubsection{Run 3 : Approche LLaMA affiné avec apprentissage par contexte}
Pour ce run, nous avons utilisé le modèle que nous appelons NER-LLaMA-3.1-8B, obtenu par fine-tuning du modèle LLaMA-3.1-8B-Instruct3 sur notre corpus synthétique décrit en section~\ref{sec2.2.1}, en utilisant la méthode LoRA avec un rang de 16. L'entraînement a été mené sur 5 époques, avec un {\it batch size} de 4, une accumulation de gradient sur 8 étapes, un taux d'apprentissage de 2e-5, et un {\it scheduler} de type cosinus.

Après affinage, le modèle est utilisé en mode {\it few-shot}, selon la méthodologie présentée en Section~\ref{sec2.2.3}. L'ajout de 10 exemples en contexte (par rapport à 0, 3 ou 5) a permis d'améliorer significativement les performances. Les exemples sont structurés sous forme de dialogues alternés \textit{user/assistant}, conformes au format d'entraînement du modèle.

\subsection{Tâche d'extraction d'événements}
Pour l'extraction d'événements, nous avons adopté, pour l'ensemble des runs, une approche unifiée fondée sur GPT-4.1 en {\it few-shot}, suivant la méthodologie du Run 1 en REN. Le prompt, présenté en Annexe~\ref{ann:B}, inclut un résumé du guide d'annotation et les 10 exemples les plus similaires (sélectionnés par similarité cosinus). Les événements sont extraits au format JSON, conforme aux attentes du défi, avec mention explicite des synonymes lorsque cela est nécessaire.

\section{Résultats}
\label{sec4}

Les résultats sur la tâche de REN, présentés dans la Table~\ref{tab:entity}, mettent en évidence des écarts significatifs de performance entre les configurations. Le modèle GPT-4.1 (run 1) affiche les meilleurs résultats, avec un score macro-F1 de 61,53~\% et un score micro-F1 de 75,79~\% sur les données de test. Ces scores traduisent une grande précision dans l'identification des entités fréquentes (micro-F1) mais aussi une capacité d'adaptation à des entités rares (macro-F1). Le modèle bénéficie probablement de la richesse contextuelle capturée à travers le mécanisme ICL ainsi que de la capacité massive de généralisation de GPT-4.1, notamment dans les domaines spécialisés, ce qui lui permet de corriger les biais de distribution.

Le modèle EvalLLM\_GLiNER (run 2) obtient des performances intermédiaires avec un score macro-F1 de 51,56~\% et un score micro-F1 de 65,22~\% sur les données de test, indiquant une bonne couverture sur les entités fréquentes, mais une efficacité modérée sur les entités plus rares. Ce profil peut être attribué à une spécialisation biomédicale du modèle GLiNER, qui améliore la précision sur les entités connues. Il est également plausible que l'intervention du LLM en post-traitement corrige certaines erreurs typiques de GLiNER, mais cette correction reste insuffisante pour atteindre la flexibilité contextuelle de GPT-4.1. Une limitation liée à la nature synthétique et peu variée des données d'affinage, affecte la robustesse face aux cas non canoniques.

Le modèle NER-LLama-3.1-8B (run 3) présente les résultats les plus faibles, avec un score macro-F1 de 40,91~\% et un score micro-F1 de 60,67~\%. Le rappel macro bas à 40,99~\% traduit une difficulté à identifier les entités peu fréquentes. Cette tendance est renforcée par la faible précision macro de 42,56~\%, indiquant un comportement erratique y compris sur des entités courantes. Malgré l'affinage supervisé sur la tâche de NER et l'utilisation de 10 exemples similaires via ICL, le modèle ne semble pas tirer pleinement parti du contexte fourni, probablement en raison de capacités de contextualisation moins performantes que celles de GPT-4.1.

Nous observons également dans les résultats une particularité notable : les performances mesurées sur le jeu de test (200 exemples) sont globalement supérieures à celles observées sur le jeu de données de développement fourni dans le cadre du challenge (40 exemples). Cela vaut notamment pour l'approche GPT-4.1 qui atteint un score macro-F1 de 61,53~\% sur les données de test contre 53,25~\% sur les données de développement, mais se vérifie aussi pour les autres runs.
Plusieurs facteurs techniques et structurels permettent d'expliquer pourquoi les performances observées sur les données de test dépassent celles obtenues sur le développement. D'abord, la taille très restreinte du jeu de données de développement (n = 40) le rend particulièrement sensible à la variance : les fluctuations statistiques sont accentuées, notamment sur des métriques comme le macro-F1, qui pénalise fortement les erreurs sur les entités rares. À l'inverse, les données de test, plus larges (n = 200), fournissent une estimation plus stable et représentative des performances.

\begin{table}[ht]
\centering
\resizebox{\textwidth}{!}{%
\begin{tabular}{|
>{\columncolor[HTML]{FFFFFF}}c 
>{\columncolor[HTML]{FFFFFF}}l 
>{\columncolor[HTML]{FFFFFF}}l 
>{\columncolor[HTML]{FFFFFF}}l 
>{\columncolor[HTML]{FFFFFF}}l 
>{\columncolor[HTML]{FFFFFF}}l 
>{\columncolor[HTML]{FFFFFF}}l 
>{\columncolor[HTML]{FFFFFF}}l 
>{\columncolor[HTML]{FFFFFF}}l 
>{\columncolor[HTML]{FFFFFF}}l 
>{\columncolor[HTML]{FFFFFF}}l 
>{\columncolor[HTML]{FFFFFF}}l 
>{\columncolor[HTML]{FFFFFF}}l |}
\hline
\multicolumn{13}{|c|}{\cellcolor[HTML]{FFFFFF}{\color[HTML]{000000} \textbf{Extraction d'Entités Nommées}}} \\ \hline

\multicolumn{1}{|l|}{\cellcolor[HTML]{FFFFFF}{\color[HTML]{000000} }} & 
\multicolumn{6}{c|}{\cellcolor[HTML]{FFFFFF}{\color[HTML]{000000} Données de développement}} & 
\multicolumn{6}{c|}{\cellcolor[HTML]{FFFFFF}{\color[HTML]{000000} Données de test}} \\ \cline{2-13}

\hline 

\multicolumn{1}{|l|}{\cellcolor[HTML]{FFFFFF}{\color[HTML]{000000} }} & 
\multicolumn{2}{c|}{\cellcolor[HTML]{FFFFFF}{\color[HTML]{000000} Précision}} & 
\multicolumn{2}{c|}{\cellcolor[HTML]{FFFFFF}{\color[HTML]{000000} Rappel}} & 
\multicolumn{2}{c|}{\cellcolor[HTML]{FFFFFF}{\color[HTML]{000000} F1}} & 
\multicolumn{2}{c|}{\cellcolor[HTML]{FFFFFF}{\color[HTML]{000000} Précision}} & 
\multicolumn{2}{c|}{\cellcolor[HTML]{FFFFFF}{\color[HTML]{000000} Rappel}} & 
\multicolumn{2}{c|}{\cellcolor[HTML]{FFFFFF}{\color[HTML]{000000} F1}} \\ \cline{2-13}

\multicolumn{1}{|l|}{\multirow{-3}{*}{\cellcolor[HTML]{FFFFFF}{\color[HTML]{000000} Run}}} & 
\multicolumn{1}{c|}{\cellcolor[HTML]{FFFFFF}{\color[HTML]{000000} Micro}} & 
\multicolumn{1}{c|}{\cellcolor[HTML]{FFFFFF}{\color[HTML]{000000} Macro}} & 
\multicolumn{1}{c|}{\cellcolor[HTML]{FFFFFF}{\color[HTML]{000000} Micro}} & 
\multicolumn{1}{c|}{\cellcolor[HTML]{FFFFFF}{\color[HTML]{000000} Macro}} & 
\multicolumn{1}{c|}{\cellcolor[HTML]{FFFFFF}{\color[HTML]{000000} Micro}} & 
\multicolumn{1}{c|}{\cellcolor[HTML]{FFFFFF}{\color[HTML]{000000} Macro}} & 
\multicolumn{1}{c|}{\cellcolor[HTML]{FFFFFF}{\color[HTML]{000000} Micro}} & 
\multicolumn{1}{c|}{\cellcolor[HTML]{FFFFFF}{\color[HTML]{000000} Macro}} & 
\multicolumn{1}{c|}{\cellcolor[HTML]{FFFFFF}{\color[HTML]{000000} Micro}} & 
\multicolumn{1}{c|}{\cellcolor[HTML]{FFFFFF}{\color[HTML]{000000} Macro}} & 
\multicolumn{1}{c|}{\cellcolor[HTML]{FFFFFF}{\color[HTML]{000000} Micro}} & 
\multicolumn{1}{c|}{\cellcolor[HTML]{FFFFFF}{\color[HTML]{000000} Macro}} \\ \hline

\multicolumn{1}{|c|}{\cellcolor[HTML]{FFFFFF}{\color[HTML]{000000} Run 1}} & 
\multicolumn{1}{l|}{\cellcolor[HTML]{FFFFFF}{\color[HTML]{000000} \textbf{64.1}}} & 
\multicolumn{1}{l|}{\cellcolor[HTML]{FFFFFF}{\color[HTML]{000000} \textbf{51.67}}} & 
\multicolumn{1}{l|}{\cellcolor[HTML]{FFFFFF}{\color[HTML]{000000} \textbf{66.34}}} & 
\multicolumn{1}{l|}{\cellcolor[HTML]{FFFFFF}{\color[HTML]{000000} \textbf{55.69}}} & 
\multicolumn{1}{l|}{\cellcolor[HTML]{FFFFFF}{\color[HTML]{000000} \textbf{65.2}}} & 
\multicolumn{1}{l|}{\cellcolor[HTML]{FFFFFF}{\color[HTML]{000000} \textbf{53.25}}} & 
\multicolumn{1}{l|}{\cellcolor[HTML]{FFFFFF}{\color[HTML]{000000} \textbf{74,18}}} & 
\multicolumn{1}{l|}{\cellcolor[HTML]{FFFFFF}{\color[HTML]{000000} \textbf{59,44}}} & 
\multicolumn{1}{l|}{\cellcolor[HTML]{FFFFFF}{\color[HTML]{000000} \textbf{77,48}}} & 
\multicolumn{1}{l|}{\cellcolor[HTML]{FFFFFF}{\color[HTML]{000000} \textbf{64,78}}} & 
\multicolumn{1}{l|}{\cellcolor[HTML]{FFFFFF}{\color[HTML]{000000} \textbf{75,79}}} & 
{\color[HTML]{000000} \textbf{61,53}} \\ \hline

\multicolumn{1}{|c|}{\cellcolor[HTML]{FFFFFF}{\color[HTML]{000000} Run 2}} & 
\multicolumn{1}{l|}{\cellcolor[HTML]{FFFFFF}{\color[HTML]{000000} 49.57}} & 
\multicolumn{1}{l|}{\cellcolor[HTML]{FFFFFF}{\color[HTML]{000000} 40.42}} & 
\multicolumn{1}{l|}{\cellcolor[HTML]{FFFFFF}{\color[HTML]{000000} 52.51}} & 
\multicolumn{1}{l|}{\cellcolor[HTML]{FFFFFF}{\color[HTML]{000000} 53.85}} & 
\multicolumn{1}{l|}{\cellcolor[HTML]{FFFFFF}{\color[HTML]{000000} 51.0}} & 
\multicolumn{1}{l|}{\cellcolor[HTML]{FFFFFF}{\color[HTML]{000000} 43.03}} & 
\multicolumn{1}{l|}{\cellcolor[HTML]{FFFFFF}{\color[HTML]{000000} 61,23}} & 
\multicolumn{1}{l|}{\cellcolor[HTML]{FFFFFF}{\color[HTML]{000000} 47,99}} & 
\multicolumn{1}{l|}{\cellcolor[HTML]{FFFFFF}{\color[HTML]{000000} 69,77}} & 
\multicolumn{1}{l|}{\cellcolor[HTML]{FFFFFF}{\color[HTML]{000000} 60,29}} & 
\multicolumn{1}{l|}{\cellcolor[HTML]{FFFFFF}{\color[HTML]{000000} 65,22}} & 
{\color[HTML]{000000} 51,56} \\ \hline

\multicolumn{1}{|c|}{\cellcolor[HTML]{FFFFFF}{\color[HTML]{000000} Run 3}} & 
\multicolumn{1}{l|}{\cellcolor[HTML]{FFFFFF}{\color[HTML]{000000} 52.74}} & 
\multicolumn{1}{l|}{\cellcolor[HTML]{FFFFFF}{\color[HTML]{000000} 36.77}} & 
\multicolumn{1}{l|}{\cellcolor[HTML]{FFFFFF}{\color[HTML]{000000} 46.63}} & 
\multicolumn{1}{l|}{\cellcolor[HTML]{FFFFFF}{\color[HTML]{000000} 33.53}} & 
\multicolumn{1}{l|}{\cellcolor[HTML]{FFFFFF}{\color[HTML]{000000} 49.5}} & 
\multicolumn{1}{l|}{\cellcolor[HTML]{FFFFFF}{\color[HTML]{000000} 34.52}} & 
\multicolumn{1}{l|}{\cellcolor[HTML]{FFFFFF}{\color[HTML]{000000} 63,44}} & 
\multicolumn{1}{l|}{\cellcolor[HTML]{FFFFFF}{\color[HTML]{000000} 42,56}} & 
\multicolumn{1}{l|}{\cellcolor[HTML]{FFFFFF}{\color[HTML]{000000} 58,14}} & 
\multicolumn{1}{l|}{\cellcolor[HTML]{FFFFFF}{\color[HTML]{000000} 40,99}} & 
\multicolumn{1}{l|}{\cellcolor[HTML]{FFFFFF}{\color[HTML]{000000} 60,67}} & 
{\color[HTML]{000000} 40,91} \\ \hline
\end{tabular}
}
\caption{Performances globales pour chaque run sur les données de développement et les données de test pour la tâche de reconnaissance d'entités nommées (REN).}
\label{tab:entity}
\end{table}

\begin{figure}[ht] 
\begin{center} 
\includegraphics[width=1\linewidth]{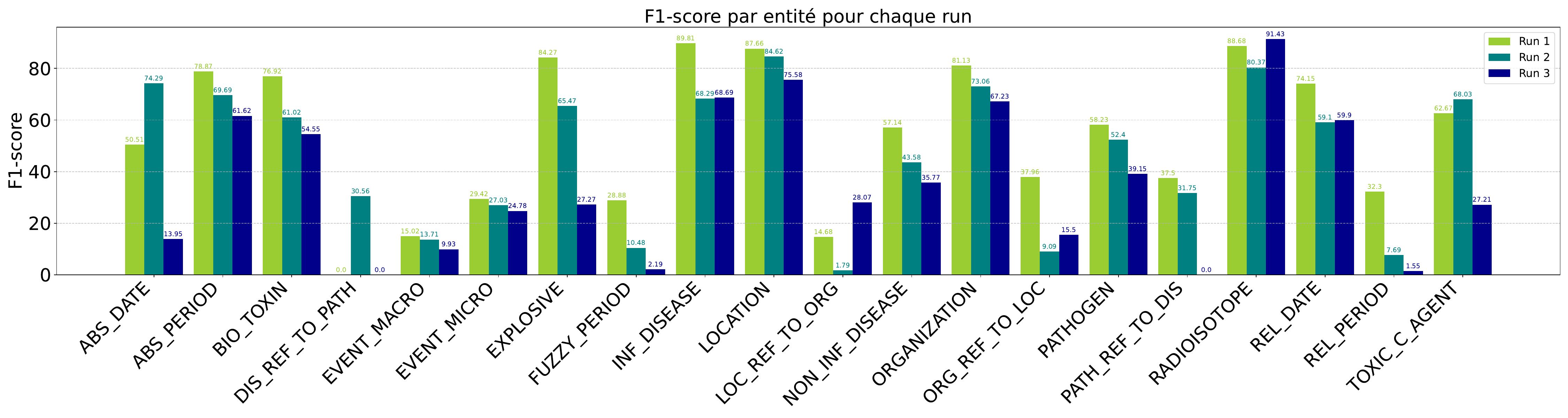}
\end{center} 
\caption{Performances en F1-score par étiquette sur le jeu de données de test, comparées entre les trois configurations expérimentales (Run 1, Run 2 et Run 3).}
\label{fig:labels_f1} \
\end{figure}

Nous pouvons observer que les scores obtenus par entité (Figure~\ref{fig:labels_f1}) reflètent en grande partie leur représentation dans les données de développement (Figure~\ref{fig:labels_dist}). Les entités les plus fréquentes dans le corpus initial, comme LOCATION (324 occurrences) ou ORGANIZATION (330), bénéficient d'une meilleure couverture en {\it few-shot}, ce qui se traduit par des scores F1 élevés (jusqu'à 87~\%). Leur forte présence dans le corpus synthétique (3~771 et 1~557 occurrences) renforce également l'impact du fine-tuning dans les runs 2 (EvalLLM\_GLiNER) et 3 (NER-LLama-3.1-8B). À l'inverse, des entités très peu présentes dans les données initiales, telles que ABS\_DATE (2 occurrences) ou DIS\_REF\_TO\_PATH (1 occurrence), restent difficiles à modéliser malgré leur augmentation, notamment en raison de la faible diversité d'exemples exploitables pour le {\it few-shot}.

Les résultats obtenus pour la tâche d'extraction d'événements, présentés dans la Table~\ref{tab:event-test}, mettent en évidence des performances globalement modestes sur le jeu de test au niveau Event (qui requiert l'identification d'un triplet structuré — élément central, lieu, date/période — dont chacun des composants doit être correctement détecté) avec des scores micro-F1 allant de 24,78~\% (Run 3) à 29,42~\% (Run 1), et des scores macro-F1 encore plus faibles, compris entre 9,93~\% et 15,02~\%. Ces performances contrastent fortement avec celles obtenues sur les données de développement, où les scores atteignent 64,82~\% en micro-F1 et 65,6~\% en macro-F1, indiquant un écart substantiel de généralisation entre les deux jeux de données. Cet écart s'explique en grande partie par le fait que les entités du corpus de développement, issues du jeu fourni par l'organisation du challenge, sont considérées comme de qualité de référence, tandis que celles du corpus de test ont été produites automatiquement par nos différents systèmes d'extraction, et peuvent donc comporter des erreurs d'identification ou de typage qui impactent la structuration des événements.

Les performances au niveau Document sur le jeu de test confirment la tendance observée au niveau Event : les scores chutent significativement par rapport aux données de développement. Le meilleur score micro-F1 atteint 46,11~\% (Run 1), contre 94,87~\% sur le corpus de développement.
L'écart significatif observé entre les performances en REN et celles en extraction d'événements au niveau Event s'explique par la complexité multi-niveaux de cette dernière. L'annotation d'un événement au niveau Event nécessite l'identification d'un triplet structuré — élément central, lieu, date/période — dont chacun des composants doit être précisément détecté et associé dans le bon contexte discursif. Cette structure impose l'extraction préalable d'entités hétérogènes selon les règles strictes du guide d'annotation.

Bien que l'approche d'extraction d'événements utilisée dans les trois runs repose sur un même modèle — GPT-4.1 en ICL avec prompting optimisé — des écarts notables sont observés entre les runs. Ces variations ne résultent pas d'une différence dans la composante événementielle elle-même, mais trouvent leur origine dans la qualité différenciée de l'extraction d'entités. En effet, un événement ne peut être détecté correctement que si ses entités constitutives sont, au préalable, extraites avec précision et contextualisation. À ce titre, le Run 1, basé sur l'exploitation directe de GPT-4.1 pour la REN avec un prompt incluant un contexte {\it few-shot} soigneusement sélectionné, a produit les entités les plus fiables (Table~\ref{tab:entity}), ce qui a mécaniquement amélioré la qualité des événements extraits. À l'inverse, les erreurs d'omission ou de segmentation dans la REN des Runs 2 et 3 ont conduit à des événements mal formés ou partiels, et donc à des performances en baisse. Cette observation souligne le caractère cumulatif et interdépendant des deux sous-tâches du défi : l'extraction des événements ne peut surpasser la qualité de la REN sous-jacente.

\begin{table}[H]
\centering
\resizebox{\textwidth}{!}{%
\begin{tabular}{|
>{\columncolor[HTML]{FFFFFF}}c 
>{\columncolor[HTML]{FFFFFF}}c 
>{\columncolor[HTML]{FFFFFF}}c 
>{\columncolor[HTML]{FFFFFF}}c 
>{\columncolor[HTML]{FFFFFF}}c 
>{\columncolor[HTML]{FFFFFF}}c 
>{\columncolor[HTML]{FFFFFF}}c 
>{\columncolor[HTML]{FFFFFF}}c 
>{\columncolor[HTML]{FFFFFF}}c 
>{\columncolor[HTML]{FFFFFF}}c 
>{\columncolor[HTML]{FFFFFF}}c 
>{\columncolor[HTML]{FFFFFF}}c 
>{\columncolor[HTML]{FFFFFF}}c |}
\hline
\multicolumn{13}{|c|}{\cellcolor[HTML]{FFFFFF}{\color[HTML]{000000} \textbf{\begin{tabular}[c]{@{}c@{}}Extraction d’Événements (Données de Test)\end{tabular}}}}                                                                                                                                                                                                                                                                                                                                                                                                                                                                                                                                                                                                                                                                                                                                                                         \\ \hline
\multicolumn{1}{|c|}{\cellcolor[HTML]{FFFFFF}}                             & \multicolumn{6}{c|}{\cellcolor[HTML]{FFFFFF}Event}                                                                                                                                                                                                                                                                                                                                                                                                                                                                         & \multicolumn{6}{c|}{\cellcolor[HTML]{FFFFFF}Document}                                                                                                                                                                                                                                                                                \\ \cline{2-13} 
\multicolumn{1}{|c|}{\cellcolor[HTML]{FFFFFF}}                             & \multicolumn{2}{c|}{\cellcolor[HTML]{FFFFFF}{\color[HTML]{000000} Précision}}                                                                                          & \multicolumn{2}{c|}{\cellcolor[HTML]{FFFFFF}{\color[HTML]{000000} Rappel}}                                                                                              & \multicolumn{2}{c|}{\cellcolor[HTML]{FFFFFF}{\color[HTML]{000000} F1}}                                                                                                  & \multicolumn{2}{c|}{\cellcolor[HTML]{FFFFFF}Précision}                                                                    & \multicolumn{2}{c|}{\cellcolor[HTML]{FFFFFF}Rappel}                                                                       & \multicolumn{2}{c|}{\cellcolor[HTML]{FFFFFF}F1}                              \\ \cline{2-13} 
\multicolumn{1}{|c|}{\multirow{-3}{*}{\cellcolor[HTML]{FFFFFF}Run}}        & \multicolumn{1}{c|}{\cellcolor[HTML]{FFFFFF}{\color[HTML]{000000} Micro}}         & \multicolumn{1}{c|}{\cellcolor[HTML]{FFFFFF}{\color[HTML]{000000} Macro}}          & \multicolumn{1}{c|}{\cellcolor[HTML]{FFFFFF}{\color[HTML]{000000} Micro}}          & \multicolumn{1}{c|}{\cellcolor[HTML]{FFFFFF}{\color[HTML]{000000} Macro}}          & \multicolumn{1}{c|}{\cellcolor[HTML]{FFFFFF}{\color[HTML]{000000} Micro}}          & \multicolumn{1}{c|}{\cellcolor[HTML]{FFFFFF}{\color[HTML]{000000} Macro}}          & \multicolumn{1}{c|}{\cellcolor[HTML]{FFFFFF}Micro}          & \multicolumn{1}{c|}{\cellcolor[HTML]{FFFFFF}Macro}          & \multicolumn{1}{c|}{\cellcolor[HTML]{FFFFFF}Micro}          & \multicolumn{1}{c|}{\cellcolor[HTML]{FFFFFF}Macro}          & \multicolumn{1}{c|}{\cellcolor[HTML]{FFFFFF}Micro}          & Macro          \\ \hline
\multicolumn{1}{|c|}{\cellcolor[HTML]{FFFFFF}{\color[HTML]{000000} Run 1}} & \multicolumn{1}{c|}{\cellcolor[HTML]{FFFFFF}{\color[HTML]{000000} 28,43}}         & \multicolumn{1}{c|}{\cellcolor[HTML]{FFFFFF}{\color[HTML]{000000} \textbf{15,02}}} & \multicolumn{1}{c|}{\cellcolor[HTML]{FFFFFF}{\color[HTML]{000000} \textbf{30,47}}} & \multicolumn{1}{c|}{\cellcolor[HTML]{FFFFFF}{\color[HTML]{000000} \textbf{16,55}}} & \multicolumn{1}{c|}{\cellcolor[HTML]{FFFFFF}{\color[HTML]{000000} \textbf{29,42}}} & \multicolumn{1}{c|}{\cellcolor[HTML]{FFFFFF}{\color[HTML]{000000} \textbf{15,02}}} & \multicolumn{1}{c|}{\cellcolor[HTML]{FFFFFF}\textbf{38,83}} & \multicolumn{1}{c|}{\cellcolor[HTML]{FFFFFF}\textbf{44,62}} & \multicolumn{1}{c|}{\cellcolor[HTML]{FFFFFF}\textbf{56,74}} & \multicolumn{1}{c|}{\cellcolor[HTML]{FFFFFF}\textbf{46,54}} & \multicolumn{1}{c|}{\cellcolor[HTML]{FFFFFF}\textbf{46,11}} & \textbf{43,74} \\ \hline
\multicolumn{1}{|c|}{\cellcolor[HTML]{FFFFFF}{\color[HTML]{000000} Run 2}} & \multicolumn{1}{c|}{\cellcolor[HTML]{FFFFFF}{\color[HTML]{000000} 26,55}}         & \multicolumn{1}{c|}{\cellcolor[HTML]{FFFFFF}{\color[HTML]{000000} 14,46}}          & \multicolumn{1}{c|}{\cellcolor[HTML]{FFFFFF}{\color[HTML]{000000} 27,53}}          & \multicolumn{1}{c|}{\cellcolor[HTML]{FFFFFF}{\color[HTML]{000000} 14,3}}           & \multicolumn{1}{c|}{\cellcolor[HTML]{FFFFFF}{\color[HTML]{000000} 27,03}}          & \multicolumn{1}{c|}{\cellcolor[HTML]{FFFFFF}{\color[HTML]{000000} 13,71}}          & \multicolumn{1}{c|}{\cellcolor[HTML]{FFFFFF}35,29}          & \multicolumn{1}{c|}{\cellcolor[HTML]{FFFFFF}40,92}          & \multicolumn{1}{c|}{\cellcolor[HTML]{FFFFFF}55,71}          & \multicolumn{1}{c|}{\cellcolor[HTML]{FFFFFF}42,23}          & \multicolumn{1}{c|}{\cellcolor[HTML]{FFFFFF}43,21}          & 40,05          \\ \hline
\multicolumn{1}{|c|}{\cellcolor[HTML]{FFFFFF}{\color[HTML]{000000} Run 3}} & \multicolumn{1}{c|}{\cellcolor[HTML]{FFFFFF}{\color[HTML]{000000} \textbf{28,5}}} & \multicolumn{1}{c|}{\cellcolor[HTML]{FFFFFF}{\color[HTML]{000000} 10,49}}          & \multicolumn{1}{c|}{\cellcolor[HTML]{FFFFFF}{\color[HTML]{000000} 21,91}}          & \multicolumn{1}{c|}{\cellcolor[HTML]{FFFFFF}{\color[HTML]{000000} 10,08}}          & \multicolumn{1}{c|}{\cellcolor[HTML]{FFFFFF}{\color[HTML]{000000} 24,78}}          & \multicolumn{1}{c|}{\cellcolor[HTML]{FFFFFF}{\color[HTML]{000000} 9,93}}           & \multicolumn{1}{c|}{\cellcolor[HTML]{FFFFFF}32,43}          & \multicolumn{1}{c|}{\cellcolor[HTML]{FFFFFF}34,35}          & \multicolumn{1}{c|}{\cellcolor[HTML]{FFFFFF}42,55}          & \multicolumn{1}{c|}{\cellcolor[HTML]{FFFFFF}33,8}           & \multicolumn{1}{c|}{\cellcolor[HTML]{FFFFFF}36,81}          & 33,23          \\ \hline
\end{tabular}
}
\caption{Performances globales pour chaque run sur les données de test pour la tâche d'extraction d'événements.}
\label{tab:event-test}
\end{table}

\section{Impact Environnemental}
\label{sec5}
Nous présentons dans la Table~\ref{tab:co2} les données concernant les coûts de traitement, les émissions de CO2 et les temps d'exécution pour chaque système selon les tâches réalisées. Il est essentiel de noter que l'empreinte carbone mesurée ici couvre uniquement l'inférence et le fine-tuning, mais exclut l'entraînement initial des modèles, étape qui génère l'impact environnemental le plus significatif. Les résultats révèlent des différences importantes entre les modèles. Pour la tâche de REN, EvalLLM-GLiNER (Run 2) présente l'empreinte carbone la plus faible en inférence (0,2g de CO2 en 1 minute), tandis que le fine-tuning de LLaMA-3.1-8B (Run 3) génère l'impact le plus élevé (91,23g de CO2 en 71 minutes). 
\begin{table}[H]
\centering
\resizebox{\columnwidth}{!}{%
\begin{tabular}{|c|cc|c|c|c|}
\hline
\textbf{Tâche} & \multicolumn{2}{c|}{\textbf{Run}} & \textbf{Temps (min)} & \textbf{CO2 (g)} & \textbf{Coût (€)} \\ \hline
\multirow{5}{*}{\textbf{REN}} & \multicolumn{1}{c|}{GPT-4.1 (Run 1)} & Inférence & 19 & 4,71 & 3,76 \\ \cline{2-6} 
 & \multicolumn{1}{c|}{\multirow{2}{*}{EvalLLM-GLiNER (Run 2)}} & Fine-tuning & 8 & 2,86 & 0,05 \\ \cline{3-6} 
 & \multicolumn{1}{c|}{} & Inférence & 1 & 0,2 & 0,006 \\ \cline{2-6} 
 & \multicolumn{1}{c|}{\multirow{2}{*}{NER-LLaMA-3.1-8B (Run 3)}} & Fine-tuning & 71 & 91,23 & 5,75 \\ \cline{3-6} 
 & \multicolumn{1}{c|}{} & Inférence & 23 & 19,7 & 0,3 \\ \hline
\multirow{3}{*}{\textbf{Extraction d’événements}} & \multicolumn{1}{c|}{GPT-4.1 (Run 1)} & \multirow{3}{*}{Inférence} & 18 & 4,45 & 4,77 \\ \cline{2-2} \cline{4-6} 
 & \multicolumn{1}{c|}{GPT-4.1 (Run 2)} &  & 17 & 4,21 & 4,69 \\ \cline{2-2} \cline{4-6} 
 & \multicolumn{1}{c|}{GPT-4.1 (Run 3)} &  & 17 & 4,21 & 4,63 \\ \hline
\textbf{Augmentation de données} & \multicolumn{1}{c|}{GPT-4.1 (Run 2 et 3)} & Inférence & 8 & 1,98 & 2,03 \\ \hline
\textbf{\begin{tabular}[c]{@{}c@{}}Vérification des sorties \\ de GLiNER\end{tabular}} & \multicolumn{1}{c|}{GPT-4.1 (Run 2)} & Inférence & 22 & 5,44 & 0,80 \\ \hline
\end{tabular}%
}
\caption{Émissions de CO2, temps d'exécution et coûts par tâche et modèle/exécution}
\label{tab:co2}
\end{table}

Il convient de noter que l'empreinte carbone élevée de NER-LLaMA-3.1-8B en inférence (19,7g de CO2) par rapport à GPT-4.1 (4,71g de CO2) s'explique principalement par les différences d'infrastructure : NER-LLaMA-3.1-8B est exécuté sur l'infrastructure Jean Zay avec une estimation basée sur la consommation énergétique directe, tandis que GPT-4.1 utilise l'API Batch d'OpenAI qui pourrait bénéficier d'optimisations d'infrastructure cloud différents. Ces résultats reposent sur des mesures CodeCarbon \footnote{\url{https://github.com/mlco2/codecarbon?tab=readme-ov-file}} pour GPT-4.1 et des estimations extraites de la documentation du supercalculateur Jean Zay\footnote{\url{https://www.edari.fr/documentation/index.php/Documentation_complète\#Suivi_du_bilan_carbone_du_projet}} pour NER-LLaMA-3.1-8B et EvalLLM-GLiNER.

\section{Discussions}
\label{sec6} 
Nous soulignons d'abord la taille restreinte du jeu de données d'entraînement, limité à 40 textes, ce qui situe nos travaux dans un cadre expérimental très contraint. Une telle limitation nuit à la performance en extraction d'événements, tâche exigeante nécessitant de nombreux exemples pour permettre aux modèles de converger et de généraliser efficacement. Le recours exclusif au guide d’annotation ne suffit, à ce jour, à garantir des performances satisfaisantes.

Des écarts importants entre les performances en REN et pour l'extraction d'événements ont été constatés. La REN est une tâche historiquement documentée dans le domaine du TAL, considérée comme plus simple contrairement à l'extraction d'événements qui fait encore l'objet de nombreux défis. Ces écarts renforcent notre hypothèse selon laquelle les LLM requièrent un volume de données conséquent pour atteindre des performances satisfaisantes, comme le suggèrent également les travaux de~\citep{piedboeuf_evaluation_2024}.
Ainsi avoir une répartition à la défaveur de l'apprentissage dans notre cas, 40 textes contre 200 contenus dans l'échantillon de test, peut expliquer en partie les résultats que nous obtenons. 

Cependant, il semble également important de nuancer notre propos car la faible quantité de données ne peut être le seul facteur explicatif. La tâche s'effectue dans le cadre de la veille sanitaire journalistique, une thématique que nous pouvons rapprocher du domaine biomédical sur la langue française. Ce domaine encore aujourd'hui peu doté~\citep{labrak:hal-04470938} pour le français, constitue une difficulté supplémentaire. Il se caractérise par l'usage de terminologies spécialisées, une forte ambiguïté sémantique de certains termes médicaux, les limitations des tokenizers actuels lorsqu'ils sont appliqués au lexique biomédical~\citep{labrak_how_2024}, ainsi qu'un manque de diversité des données disponibles, une partie non négligeable des corpus étant issue de traductions de jeux de données libres initialement rédigés en anglais.

Nous considérons que la combinaison de ces deux facteurs constitue un premier élément de réponse, susceptible d'améliorer les performances des modèles, une fois ces problèmes résolus.

\section{Conclusion}
\label{sec7}
Les résultats obtenus dans le cadre de ce défi confirment l'importance du prompt engineering et de l'apprentissage par contexte (ICL) pour la reconnaissance d'entités nommées (REN), où GPT-4.1 surpasse nettement les modèles affinés sur données synthétiques. L'ajout de données générées améliore les performances pour certaines entités rares, mais cet effet reste limité à la tâche de REN. Pour l'extraction d'événements, qui repose exclusivement sur GPT-4.1 en ICL dans nos trois configurations, les performances restent modestes, notamment en raison du faible nombre d'exemples disponibles dans le corpus initial pour la sélection {\it few-shot}. Ces résultats soulignent que, dans un cadre à faibles ressources, la qualité du prompt engineering et des exemples de démonstration joue un rôle déterminant, et qu'en l'absence d'un fine-tuning dédié, les modèles pré-entraînés peuvent montrer des limites sur des tâches complexes et structurées comme l'extraction d'événements.

\bibliographystyle{coria-taln2025}
\bibliography{biblio}

\newpage
\appendix
\section{Prompt-système pour la tâche de REN fourni à GPT-4.1}
\label{ann:A}
\begin{figure}[H]
\centering
\begin{tcolorbox}[
    colback=gray!10,
    colframe=blue!60,
    title={\textbf{SYSTEM\_PROMPT}},
    fonttitle=\bfseries,
    left=3pt,
    right=3pt,
    top=3pt,
    bottom=3pt,
    breakable
]
\scriptsize
\textbf{You are an expert French medical annotator.}

\vspace{0.2cm}
\textbf{====== TASK ======}\\
1. Read the input French text.\\
2. Insert XML tags \textbf{in-line} around every entity mention according to the label definitions provided.\\
\quad → Example: Le \texttt{<INF\_DISEASE>paludisme</INF\_DISEASE>} est endémique.\\
3. Return \textbf{only} the full annotated text. No commentary, no metadata.

→ Think step by step \textbf{internally}, but output only the final tagged text.

\vspace{0.2cm}
\textbf{======= ANNOTATION RULES ========}\\
• Use only the labels from the glossary.\\
• Exclude determiners, pronouns, and punctuation from entity spans.\\
• If an entity is \textbf{discontinuous}, tag \textbf{each contiguous part separately} with the same label and shared \texttt{ent\_id}.\\
\quad → Ex: les \texttt{<PATHOGEN ent\_id="P1"><PATHOGEN ent\_id="P2">virus</PATHOGEN></PATHOGEN>} de la \texttt{<PATHOGEN ent\_id="P1">dengue</PATHOGEN>} et du \texttt{<PATHOGEN ent\_id="P2">chikungunya</PATHOGEN>}\\
\quad \texttt{<ORGANIZATION ent\_id="O1">Agence régionale de santé</ORGANIZATION>} (\texttt{<ORGANIZATION ent\_id="O2">ARS</ORGANIZATION>}) \texttt{<ORGANIZATION ent\_id="O1"><ORGANIZATION ent\_id="O2">d'Île de France</ORGANIZATION></ORGANIZATION>}\\
• Tags must not cross paragraph boundaries.\\
• Ignore misspellings, generic terms ("virus", "bactérie", etc.), and pronouns.\\
• Do \textbf{not} generate any tag that does not exist in the input.\\
• Use valid XML syntax. Tags must be correctly opened/closed and perfectly nested.\\
• Overlapping tags are allowed \textbf{only} for discontinuous spans (as shown above).

\vspace{0.2cm}
\textbf{======= LABEL GLOSSARY =======}\\
\ding{51} = tag it \ding{55} = don't tag it

→ \textbf{Document-level metadata}\\
• \texttt{DOC\_AUTHOR} \ding{51} "Jean Dupont" (byline only) \ding{55} in body\\
• \texttt{DOC\_SOURCE} \ding{51} "AFP", "Reuters" \ding{55} "la presse"

→ \textbf{Diseases \& Pathogens}\\
• \texttt{INF\_DISEASE} \ding{51} grippe, rougeole \ding{55} "maladie", "infection"\\
• \texttt{NON\_INF\_DISEASE} \ding{51} cancer, diabète \ding{55} syndromes mixtes\\
• \texttt{PATHOGEN} \ding{51} Escherichia coli, virus Ebola \ding{55} "virus" (generic)\\
• \texttt{DIS\_REF\_TO\_PATH} \ding{51} paludisme in "parasites tels que le paludisme" \ding{55} paludisme as disease\\
• \texttt{PATH\_REF\_TO\_DIS} \ding{51} VIH in "cas de VIH" \ding{55} virus VIH

→ \textbf{Toxins, Chemicals, Explosives}\\
• \texttt{RADIOISOTOPE} \ding{51} uranium 238, césium-137\\
• \texttt{TOXIC\_C\_AGENT} \ding{51} sarin, chlore gazeux\\
• \texttt{EXPLOSIVE} \ding{51} TNT, RDX\\
• \texttt{BIO\_TOXIN} \ding{51} ricine, toxine botulique

→ \textbf{Locations \& Organizations}\\
• \texttt{LOCATION} \ding{51} Paris, Rhône, Alpes \ding{55} pronouns, "le pays"\\
• \texttt{ORGANIZATION} \ding{51} OMS, hôpital Georges-Pompidou\\
• \texttt{LOC\_REF\_TO\_ORG} \ding{51} Paris (dans "Paris annonce…")\\
• \texttt{ORG\_REF\_TO\_LOC} \ding{51} centrale nucléaire de Tchernobyl

→ \textbf{Dates \& Time References}\\
• \texttt{ABS\_DATE} \ding{51} 8 janvier 2025, 01/08/2025\\
• \texttt{REL\_DATE} \ding{51} hier, lundi dernier, 8 janvier (sans année)\\
• \texttt{DOC\_DATE} \ding{51} date en tête d'article\\
• \texttt{ABS\_PERIOD} \ding{51} mars 2024, du 1er au 3 mai 2024\\
• \texttt{REL\_PERIOD} \ding{51} la semaine dernière, du 10 au 20 mai\\
• \texttt{FUZZY\_PERIOD} \ding{51} ces dernières années, depuis plusieurs semaines

\vspace{0.2cm}
\textbf{====== CONSTRAINTS ======}\\
1. Output must contain \textbf{valid XML} with correct nesting.\\
2. A token may belong to multiple tags \textbf{only} when discontinuity requires it.\\
3. Never output tags for absent entities or unsupported labels.

\vspace{0.2cm}
\textbf{====== EXAMPLES ======}\\
\textit{[Examples would follow here]}
\end{tcolorbox}
\label{prompt:system}
\end{figure}

\section{Prompt-système pour la tâche d'extraction d'événements fourni à GPT-4.1}
\label{ann:B}
\begin{figure}[H]
\centering
\begin{tcolorbox}[
    colback=gray!10,
    colframe=orange!70,
    title={\textbf{SYSTEM\_PROMPT}},
    fonttitle=\bfseries,
    left=3pt,
    right=3pt,
    top=3pt,
    bottom=3pt,
]
\scriptsize
\textbf{You are an epidemiology analyst. Your job is to extract structured events from French articles.}
\vspace{0.3cm}
\textbf{========== TASK ==========}\\
\textbf{INPUT:}\\
• A French article.\\
• A list of extracted named entities (ID, text span, and type).\\
\textbf{OUTPUT:}\\
Return a JSON array named events following this schema:

\begin{verbatim}
[
  [
    {"attribute":"evt:central_element", "occurrences":["ID_c1", "ID_c2", ...]},
    {"attribute":"evt:associated_element", "occurrences":["ID_a1", "ID_a2", ...]}
  ],
]
\end{verbatim}

\vspace{0.2cm}
\textbf{========== RULES ==========}\\
\textbf{1. CENTRAL ELEMENT — REQUIRED (1 per event)}\\
- Must be exactly one of: INF\_DISEASE, NON\_INF\_DISEASE, PATHOGEN, DIS\_REF\_TO\_PATH, PATH\_REF\_TO\_DIS, RADIOISOTOPE, TOXIC\_C\_AGENT, EXPLOSIVE, BIO\_TOXIN\\
- Each event has exactly one central element (but it may have several synonymous IDs see Rule 4).

\textbf{2. ASSOCIATED ELEMENTS — REQUIRED (at least one location + at least one date/periode)}\\
Add all entity IDs relevant to:\\
- Locations: LOCATION, LOC\_REF\_TO\_ORG, ORG\_REF\_TO\_LOC\\
- Dates: ABS\_DATE, REL\_DATE, ABS\_PERIOD, REL\_PERIOD, FUZZY\_PERIOD, DOC\_DATE\\
- Use DOC\_DATE only if no other date is found.\\
- Prefer absolute over relative dates if both exist.

\textbf{3. WINDOW OF RELEVANCE}\\
- Start from the sentence containing the central element.\\
- If no associated location/date is there, check the adjacent sentences.

\textbf{4. SYNONYMS}\\
If several entity IDs refer to the same real‑world object (e.g. three mentions of "uranium 238", or "Paris" vs "Ville‑Lumière", or different surface forms of the same date), include all those IDs together in the same occurrences list.

\textbf{5. EVENT LIMIT}\\
- Max 10 events.\\
- If more are present, keep the 10 most relevant to public health risk.

\textbf{6. VALIDITY}\\
- Each entity ID appears in only one event.\\
- Output must be valid JSON and contain nothing else.

\vspace{0.2cm}
\textbf{========== TIPS ==========}\\
For event splitting, use this rule:\\
– Same central + coherent dates/places → merge into one event.\\
– Distant in time/space or different causes → separate events.

When in doubt between including or skipping an associated element: include it if it helps answer: Where? When? What agent?

\vspace{0.2cm}
\textbf{========== EXAMPLES ==========}\\
\textit{[Examples would follow here]}
\end{tcolorbox}
\label{prompt:epidemiology}
\end{figure}

\section{Prompt-système pour la tâche de vérification de sorties de GLiNER (Run 2) fourni à GPT-4.1}
\label{annC}
\begin{figure}[ht]
\centering
\begin{tcolorbox}[
    colback=gray!10,
    colframe=purple!50,
    title={\textbf{SYSTEM\_PROMPT}},
    fonttitle=\bfseries,
    left=3pt,
    right=3pt,
    top=3pt,
    bottom=3pt,
]
\scriptsize
\textbf{You are a biomedical named entity recognition (NER) expert. Your task is to review, correct, and complete the entity annotations in the following text using inline XML-style tags.}

\vspace{0.1cm}
\textbf{Instructions:}

• The input text already contains XML-style tags (e.g., \texttt{<RADIOISOTOPE>uranium 238</RADIOISOTOPE>}).\\
• Verify each existing tag:\\
\quad • Ensure the entity label is correct.\\
\quad • Correct any mislabeling.\\
• Tag any missing entities using only the valid labels from the glossary below.\\
• Return only the corrected and fully tagged version of the text in valid XML format — no extra text or explanation.

\vspace{0.1cm}
\textbf{Annotation Rules:}

• Use only labels from the glossary below.\\
• Exclude determiners, pronouns, and punctuation from inside tags.\\
• Tags must not cross paragraph boundaries.\\
• Do not tag generic terms like "virus", "bactérie", or any pronouns.\\
• Do not invent or use tags that are not present in the glossary below.\\
• Ensure all XML is valid: tags must be correctly opened and closed.

\vspace{0.1cm}
\textbf{Glossary of Valid Entity Labels and Definitions:}

• \texttt{<DOC\_AUTHOR>} — Document author(s).\\
• \texttt{<DOC\_SOURCE>} — The source or publisher of the document (e.g., 'AFP', 'Reuters').\\
• \texttt{<INF\_DISEASE>} — Infectious diseases (caused by bacteria, viruses, fungi, parasites, etc.).\\
• \texttt{<NON\_INF\_DISEASE>} — Non-infectious diseases (e.g., diabetes, cancer).\\
• \texttt{<PATHOGEN>} — The infectious agent itself (bacterium, virus, parasite, etc.).\\
• \texttt{<DIS\_REF\_TO\_PATH>} — A disease name used to refer to the pathogen.\\
• \texttt{<PATH\_REF\_TO\_DIS>} — A pathogen name used to refer to the disease.\\
• \texttt{<RADIOISOTOPE>} — A radioactive form of an element (e.g., polonium, uranium-238).\\
• \texttt{<TOXIC\_C\_AGENT>} — Inorganic toxic chemicals (e.g., chlorine gas).\\
• \texttt{<EXPLOSIVE>} — Any explosive substance or compound.\\
• \texttt{<BIO\_TOXIN>} — Organic chemical toxins from biological sources (e.g., ricin, botulinum toxin).\\
• \texttt{<LOCATION>} — Named geographic places (countries, cities, rivers, etc.).\\
• \texttt{<ORGANIZATION>} — Institutions or agencies with social/legal identity (e.g., WHO, Institut Pasteur).\\
• \texttt{<LOC\_REF\_TO\_ORG>} — Place name used to refer to an organization.\\
• \texttt{<ORG\_REF\_TO\_LOC>} — Organization name used to refer to the place it is located.\\
• \texttt{<ABS\_DATE>} — Exact date (e.g., "15 mars 2020").\\
• \texttt{<REL\_DATE>} — Relative date (e.g., "hier", "lundi dernier").\\
• \texttt{<DOC\_DATE>} — Document publication date.\\
• \texttt{<ABS\_PERIOD>} — Exact period (e.g., "mars 2020", "du 1er au 3 mai").\\
• \texttt{<REL\_PERIOD>} — Relative period (e.g., "les 3 derniers jours").\\
• \texttt{<FUZZY\_PERIOD>} — Vague time period (e.g., "ces dernières années", "depuis plusieurs mois").

\vspace{0.1cm}
\textbf{Examples:}

\textbf{Input:}
"La réunion a eu lieu le 12 avril 2020."\\
→ \textbf{Correction:}
"La réunion a eu lieu le \texttt{<ABS\_DATE>12 avril 2020</ABS\_DATE>}."

\textbf{Input:}
"Ces dernières années, les cas ont augmenté."\\
→ \textbf{Correction:}
"\texttt{<FUZZY\_PERIOD>Ces dernières années</FUZZY\_PERIOD>}, les cas ont augmenté."

\textbf{Input:}
"\texttt{<LOCATION>Paris</LOCATION>} a annoncé un plan d'urgence sanitaire."\\
→ \textbf{Correction:}
"\texttt{<LOC\_REF\_TO\_ORG>Paris</LOC\_REF\_TO\_ORG>} a annoncé un plan d'urgence sanitaire."

\textbf{Input:}
"Les tests ont été menés entre mars et juin 2021."\\
→ \textbf{Correction:}
"Les tests ont été menés entre \texttt{<ABS\_PERIOD>mars et juin 2021</ABS\_PERIOD>}."

\textbf{Input:}
"Le \texttt{<PATHOGEN>virus</PATHOGEN>} peut causer des dommages importants."\\
→ \textbf{Correction:}
"Le virus peut causer des dommages importants." // Do not tag generic terms like 'virus' when unspecific.

\textbf{Input:}
"Un accident a eu lieu dans la centrale nucléaire de \texttt{<LOCATION>Tchernobyl<LOCATION>}."\\
→ \textbf{Correction:}
"Un accident a eu lieu dans la \texttt{<ORG\_REF\_TO\_LOC>centrale nucléaire de Tchernobyl</ORG\_REF\_TO\_LOC>}."

\textbf{Input:}
"Le \texttt{<PATHOGEN>paludisme</PATHOGEN>} est causé par un parasite."\\
→ \textbf{Correction:}
"\texttt{<DIS\_REF\_TO\_PATH>paludisme</DIS\_REF\_TO\_PATH>} est causé par un parasite."

\textbf{Input:}
"Le \texttt{<PATHOGEN>VIH</PATHOGEN>} est une infection virale chronique."\\
→ \textbf{Correction:}
"\texttt{<PATH\_REF\_TO\_DIS>VIH</PATH\_REF\_TO\_DIS>} est une infection virale chronique."

\vspace{0.1cm}
\textbf{Only output the corrected and completed XML-tagged version of the text. Do not include any additional explanation.}
\end{tcolorbox}
\end{figure}

\end{document}